\documentclass[10pt,twocolumn,a4paper]{article}

\usepackage{cvpr}
\usepackage{times}
\usepackage{epsfig}
\usepackage{graphicx}
\usepackage{amsmath}
\usepackage{amssymb}
\usepackage{makecell}

\usepackage{{adjustbox}}

\usepackage{multirow}

\usepackage[pagebackref=true,breaklinks=true,letterpaper=true,colorlinks,bookmarks=false]{hyperref}

\cvprfinalcopy 

\def\cvprPaperID{****} 

\ifcvprfinal\fi
\begin{document}

\title{The Focus-Aspect-Polarity Model \\
for Predicting Subjective Noun Attributes in Images}

\author{Tushar Karayil$^1$\\
DFKI, Germany\\
{\tt\small tushar.karayil@dfki.de}
\and
Philipp Blandfort$^1$\\
DFKI and TUK, Germany\\
{\tt\small philipp.blandfort@dfki.de}
\and
J\"orn Hees\\
DFKI, Germany\\
{\tt\small joern.hees@dfki.de}
\and
Andreas Dengel\\
DFKI, Germany\\
{\tt\small andreas.dengel@dfki.de}
}

\maketitle

\footnotetext[1]{Equal contribution.} 

\begin{abstract}
   Subjective visual interpretation is a challenging yet important topic in computer vision.
   Many approaches reduce this problem to the prediction of adjective- or attribute-labels from images.
   However, most of these do not take attribute semantics into account, or only process the image in a holistic manner.
   Furthermore, there is a lack of relevant datasets with fine-grained subjective labels. 
   In this paper, we propose the Focus-Aspect-Polarity model to structure the process of capturing subjectivity in image processing, and introduce a novel dataset following this way of modeling.
   We run experiments on this dataset to compare several deep learning methods and find that incorporating context information based on tensor multiplication in several cases outperforms the default way of information fusion (concatenation).
\end{abstract}

\section{Introduction}
Subjectivity is the phenomenon wherein human perception is influenced by personal feelings, tastes, opinions etc. 
The variance which arises as a result of this phenomenon plays a crucial role in the visual domain.
For example, the meaning that we infer from an image can depend on: our internal templates about the stimuli \cite{stimuli},  expectations and learned biases about the visual object \cite{expectation}, context / prior visual input \cite{context}, random neural fluctuations in cortex \cite{random} and other factors like personality of the interpreting individual.
This innate diversity in interpretation has made evaluation and computational modeling of subjectivity a difficult task.

\begin{figure}[th]
\begin{center}
   \includegraphics[width=1\linewidth]{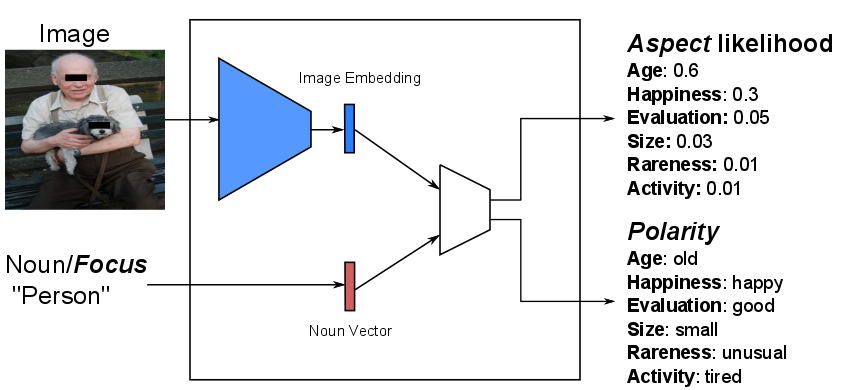}
\end{center}
   \caption{Illustration of the task. The model takes an image and a noun (\textit{focus}) present in the image as input. It outputs the corresponding \textit{aspects} and \textit{polarity} (of each \textit{aspect}). For the given image in the illustration, since the \textit{focus} is on the noun ``person'', the model identifies the \textit{aspects} ``Age'' and ``Happiness'' as the most appropriate. The \textit{polarity} provided for each \textit{aspect} determines which set of attributes suits the noun in the context of the image. For the \textit{aspect} age in the given example, the \textit{polarity} output indicates that suitable attributes for the person in the image would be ``old'', ``elderly'', ``mature'' or ``senior'', in contrast to ``young''. Attributions of images, ``Old man and his dog'' by Katie Cook, used under CC BY-NC-ND 2.0.
   }
\label{fig:modeling}
\end{figure}

The challenge in modeling subjectivity arises from two main sources.
First, subjective interpretation by definition is arbitrary in a certain sense, since there is no a priori objective taste, feeling, or opinion, and at times such context information might not be accessible at all.
In particular, this poses challenges to evaluation, and in many cases it is reasonable to expect that there will be a larger margin to a perfect score. 
Second, subjectivity tends to be more fine-grained than objectivity.
For example, in images, objectivity that is detected typically is about which entities are visible,
while subjective information is rather about characterizing how these entities or the picture as a whole differ from some expectation \cite{context, expectation}.

To attenuate these issues, previous methods typically consider holistic aspects of subjectivity (e.g. in visual sentiment analysis) or mix subjective components with non-subjective components (as in adjective-noun pairs) \cite{borth2013sentibank}.
A problem with the latter is that, in evaluation, these components are mixed and might be hard to separate later on, while the original interest was to focus on subjective parts. Additionally, existing works which use this approach do not include any sophisticated structuring of the subjective components.
We also found that there is a clear lack of datasets with more fine-grained or structured subjective aspects annotated.

In order to overcome these shortcomings, we propose a novel dataset (\textit{aspects-DB}) and the Focus-Aspect-Polarity model for subjective visual interpretation, disentangling three components of subjectivity:
1) focus: the center of attention,
2) aspect: which dimension to evaluate on, and
3) polarity: result of this evaluation.
Our proposed way of modeling is illustrated in Figure~\ref{fig:modeling}.
Briefly put, our model works as follows:
Given an image and, as context, a noun present in the image (as proxy to describe which part of the image is attended to), we would like to first identify which dimension of evaluation (represented by \textit{aspects}) one is likely use for describing the noun in the given image, and secondly predict how the noun would be evaluated with respect to these dimensions of evaluation (represented by \textit{aspect polarities}).
Finally, in this paper we analyze several methods for the emerging tasks \textit{aspect prediction} and \textit{polarity detection},
thereby providing an overview of different ways of using context information in this particular case and revealing general open issues.

The rest of the paper is organized as follows. Section~\ref{sec:relatedwork} surveys related work relevant to this paper. Section~\ref{sec:modeldataset} introduces the model and the new dataset. Section~\ref{sec:tasks} explains the two tasks that form the core of this method.  Section~\ref{sec:methods} gives a detailed description of the experiments and architectures used.  Section~\ref{sec:experiments} provides our insights and findings from the experiments along with the open questions. Section~\ref{sec:conclusion} concludes the paper with a summary and future work.

\section{Related Work}\label{sec:relatedwork}
This section can be broadly divided into three segments:
First, the methods which attempt to capture subjectivity.
Second, the methods which use Adjective-Noun pairs for this purpose.
Finally, the available attribute detection datasets.

\subsection{Detecting Subjectivity}
There have been many promising approaches which researchers have employed at detecting subjective parts of visual interpretation.
While some works focused on attributes to enhance the quality of nouns \cite{farhadi2009describing, krishna2017visual,borth2013large},
others focused on understanding the aesthetics \cite{moorthy2010towards,dhar2011high}.

\cite{borth2013large, borth2013sentibank} proposed the large scale visual sentiment ontology to detect adjective-noun pairs inside an image.
Given an image they propose to find a suitable adjective-noun pair to best describe an image from a set of adjective-noun pairs.
Although adjective noun pairs capture the sentiment to an extent, they do not reveal the degree to which this sentiment applies. Moreover, relying on  a single adjective-noun pair to describe the whole image would mean only the most prominent noun is focused upon. 

The authors of \cite{Lazaridou2015FromVA} propose a cross-modal mapping from a visual semantic space onto a linguistic space in order to automatically annotate images with adjectives.
The mapping is performed by a projection function that maps the vector representation of an image tagged with an object / attribute onto the linguistic representation of the object / attribute word. This mapping function can then be applied to any given image to obtain its linguistic projection. 
The main advantage, as claimed by \cite{Lazaridou2015FromVA}, is that of zero-shot learning, i.e., unseen attributes (not present in training) can be predicted.
However, in this approach the whole image is mapped onto an adjective without focusing on any particular noun or aspect.
Our method, on the other hand focuses on finding a suitable set of adjectives for a given noun in context of an image,
and still allows for zero-shot learning.

\subsection{Detecting Adjectives-Nouns}

Our work mainly builds on a line of work originating from the Visual Sentiment Ontology \cite{VSO} proposed by Borth et al.,
which aims at detecting adjective-noun combinations from images.
So far, the best performing method within this direction are cross-residual networks (XResNet) \cite{xresnet},
which we include in our experiments and will describe in detail in Section~\ref{sec:xresnet}.
For any given image, XResNet outputs scores for adjective-noun combinations as well as scores for all individual adjectives and nouns separately.
This means that it separates the more subjective parts of interpretation (represented by the adjectives) from the more objective (represented by the nouns).

There are two major datasets that have been used for training the above-mentioned architectures:
The Visual Sentiment Ontology (VSO) \cite{VSO} and the Multilingual Visual Sentiment Ontology (MVSO) \cite{MVSO}.
These datasets have been created from the popular photo-sharing platform Flickr.
However, the data in these cases suffers from a clear bias towards the positive attributes / adjectives \cite{kalkowski2015real}. 
\cite{xresnet} has taken some efforts for achieving a better overall balance, but even there,
for any given noun the number of associated adjectives is typically very small and the distribution heavily skewed.
More importantly, the ``feasible'' adjectives for a given noun are in most cases not mutually exclusive, at times even similar in meaning (e.g. ``smiling person'' and ``happy person''),
and yet any non-ground-truth adjective is typically considered to be wrong.
This makes it harder to interpret performances on these datasets in terms of ability to capture subjective aspects.

These issues can be overcome by considering the problem of attribute prediction to focus on the subjective part of visual interpretation:
Given an image and an entity (in our case represented by a noun) in the image, estimate the suitability of attributes
under consideration of their semantic relations. 
In this regard we created a dataset with structured and properly balanced attributes for any given noun,
thus addressing the issue of balance which was found lacking in the existing VSO and MVSO.


\subsection{Attribute datasets}
There are several popular attribute datasets available for computer vision research.

The Visual Genome \cite{visualgenome} contains over 100,000 images with fine-grained annotations, including region descriptions, object instances and visual attributes in the order of Millions.
However, the attributes in this dataset mostly relate to objective information.
Hence, most common attributes are colors like white, blue red, black 
and despite the large number of total annotations in Visual Genome, we found the number of subjective attribute instances to be too low for our purpose.

aPascal and aYahoo \cite{5206772} are two attribute datasets containing natural object-based images with attribute annotations.
Here again, the included attributes correspond to objective features, such as parts of a face like eyes, nose and so on,
which deems it inappropriate for analyzing subjective interpretation.

Another attribute dataset is the SUN Attribute Dataset \cite{patterson2014sun}, 
which contains scene attributes of the four categories ``functions / affordances'' (e.g. ``diving'', ``climbing''), ``materials'', ``surface properties'' and ``spatial envelope''.
The former three categories are restricted to objective information, and while there are several subjective attributes (such as ``scary'' or ``stressful'') in the ``spatial envelope'' category, all of these annotations are describing the scene in a holistic manner.

Overall, none of the available datasets is appropriate for focusing on more fine-grained subjective visual interpretation.

\begin{table*}[ht]
\begin{center}
    \begin{tabular}{|c|c|c|c|}
   
   \hline
   \thead{Aspect} & \thead{Noun} & \thead{Left Polarity}  & \thead{Right Polarity}\\
    \hline
   age  &   
     people &
    \begin{minipage}{.35\textwidth}
       \begin{center} 
       ``young'' \\
       \includegraphics[width=.75\linewidth, height=.4\linewidth]{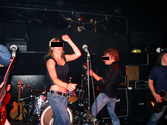} \end{center}
    \end{minipage} &
    \begin{minipage}{.35\textwidth}
          \begin{center} 
          \vspace{1ex}
          ``old'', ``elderly'', ``mature'', ``senior'', ``aged'' \\
          \includegraphics[width=.75\linewidth, height=.4\linewidth]{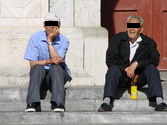} \end{center}
    \end{minipage} 
    \\[11ex]
    \hline

   activity &   
     city &
    \begin{minipage}{.35\textwidth}
        \begin{center}
        \vspace{1ex}
        ``active'', ``busy''\\
        \includegraphics[width=.75\linewidth, height=.4\linewidth]{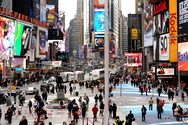} \end{center}
    \end{minipage} &
    \begin{minipage}{.35\textwidth}
         \begin{center}
         \vspace{1ex}
         ``sleepy'', ``sleeping''\\
         \includegraphics[width=.75\linewidth, height=.4\linewidth]{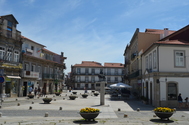} \end{center}
    \end{minipage} 
    \\[10ex]
    \hline
    
    happiness &   
     boy &
    \begin{minipage}{.35\textwidth}
         \begin{center}
         \vspace{1ex}
         ``smiling'', ``laughing'', ``happy''\\
         \includegraphics[width=.75\linewidth, height=.4\linewidth]{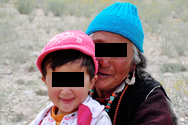} \end{center}
    \end{minipage} &
    \begin{minipage}{.35\textwidth}
       \begin{center}
       \vspace{1ex}
       ``crying'', ``sad'' \\
       \includegraphics[width=.75\linewidth, height=.4\linewidth]{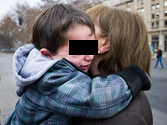} \end{center}
    \end{minipage} 
    \\[10ex]
    \hline
 
 \end{tabular} 
 \caption{Examples of ground truth data in the proposed \textit{aspects-DB} dataset. Each row represents an aspect, an example noun and a sample image which corresponds to the left and right polarities.
 Attributions of images, from left to right: ``Young Heart Attack'' by Richard Child, used under CC BY-NC-ND 2.0, ``Old Chinese Men'' by Michael Goodine, used under CC BY-NC-ND 2.0., ``Busy Times Square'' by Jim Larrison, used under CC BY-NC-ND 2.0.,  ``A quiet Saturday morning'' by Pedro, used under CC BY-NC-ND 2.0, ``Grandma looks happy'' by Praveen, used under CC BY-NC-ND 2.0, ``Boy crying'' by Francisco Osorio, used under CC BY-NC-ND 2.0}
    \label{tab:samples}
 \end{center}
\end{table*}

\section{Modeling and Dataset} \label{sec:modeldataset}

\subsection{Focus-Aspect-Polarity model for subjective visual interpretation}

The work of Borth et al. \cite{VSO} shows that adjective noun combinations are often visible and reasonably simple to automatically detect in images,
presumably because of how they contain both subjective (in the adjective) and objective (in the noun) information.

If we consider the semantics of adjectives as described by Baroni and Zamparelli in \cite{Baroni:2010:NVA:1870658.1870773},
where adjectives are interpreted as modifiers of nouns, we see that visually detecting adjective noun combinations can be understood as  
a model that combines attention and evaluation, where the noun describes where the viewer is focusing when interpreting the image and the adjective contains the subjective evaluation of this part of the image. 

For the adjective, we want to take a step further and acknowledge the fact that adjectives for the same noun are often semantically related.
In other words, they can be organized along various dimensions of evaluation.
Example for such dimensions would be size, age, cuteness or temperature.

So instead of considering any non-ground-truth adjective as wrong and thereby largely ignoring semantic relations between adjectives,
we organized the adjectives into opposing lists. 
Arranging in this manner paves way for a more appropriate evaluation as opposing adjectives (mutually exclusive) cannot occur together for the same noun.
For example, if we consider the opposing adjectives in [``cute'', ``adorable''] vs [``scary'', ``ugly''],
a classification of ``cute'' or ``adorable'' of a puppy are semantically similar,  but ``cute'' and ``scary'' cannot apply to the same puppy.  
To further elaborate, we arranged a list of mutually opposing adjectives, where each opposing list reflects a certain dimension of evaluation, which we call \textit{``aspect''}, of the noun.
The \textit{aspects} and the adjectives pertaining to these aspects that are incorporated in our dataset are listed in Table~\ref{tab:aspects} and will be derived in Section~\ref{sec:dataset}.

In summary, we separate three potential sources of subjectivity in our model:
\begin{enumerate}
  \item \textit{Focus}: Given a single image, there are typically different components one can pay attention to. For this paper we will assume that this place of focus can be captured by a noun. Note that nouns can relate to an entity in the image (such as ``dog'' or ``dude''), but also refer to the whole scene (as in ``place'') or the picture itself (``shot'').
  \item \textit{Aspect}: Once the focus has been determined, there are several potential dimensions for evaluation. For example, people in the image can be evaluated with respect to their physical size, age, level of activity and so on. In our dataset, selecting an aspect for evaluation is essentially about choosing a set of semantically related attributes.
  \item \textit{Polarity}: In our case, we chose all aspects to be represented by mutually exclusive sets of adjectives, such that evaluating each aspect amounts to a binary decision problem. For example, physical size would have adjectives like ``small'', ``tiny'', ``short'' on one side and ``tall'', ``big'', ``huge'' on the other. Picking a certain polarity then means to say that one of the adjectives from this side is appropriate to be used as an attribute for the given noun.
\end{enumerate}

The following points summarize the key features of this method of modeling:
\begin{itemize}
  \item Three different sources of subjectivity are disentangled. This brings about the possibility to evaluate these components separately, and can potentially be exploited by computational models, e.g. by learning biases at these distinct levels. 
  \item Semantic relations between attributes are respected. In particular, by detecting aspect polarities instead of individual attributes, we treat attributes of the same polarity as being synonymous for the given aspect. We thereby avoid to consider any attribute as wrong if it means the same but is merely phrased differently, as it is for example done when using adjective noun combinations or single attributes as independent class labels.
  \item This modeling leads to a more sensible way of 0-shot learning for attribute detection, i.e., predicting subjective attributes for nouns for which they were not available during training time. We will explore this direction below.
\end{itemize}

\subsection{Compiling the dataset} \label{sec:dataset}
To overcome the shortcomings of the existing datasets mentioned above, and to have a fair evaluation for experiments,
we decided to create a new dataset called \textit{aspects-DB} for subjective visual interpretation, following the AAP model.
We will now describe the steps we took for building the dataset.

First, based on the Visual Sentiment Ontology, we compiled thematic lists of nouns,
focusing on terms from the urban environment since such images often contain several entities: 
\begin{itemize}
\item \textit{people}: ``person'', ``people'', ``guy'', ``girl'', ``woman'', ``man'', ``baby'', ``boy'', ``child''
\item \textit{animals}: ``cat'', ``dog'', ``animal'', ``pet'', ``puppy'', ``kitten'', ``bird''
\item \textit{buildings}: ``building'', ``house'', ``architecture'', ``hotel'', ``church'', ``restaurant''
\item \textit{scene}: ``street'', ``place'', ``view'', ``city'', ``event'', ``neighborhood'', ``location''
\item \textit{plants}: ``tree'', ``plant'', ``flower''
\end{itemize}

Second, based on semantic adjective classes in GermaNet \cite{germanet} 
we selected a list of adjective classes that can apply to these nouns and can be visible in images.
Examples for such semantic classes are \textit{appearance} (``pretty'', ``ugly'', ...), \textit{size} (``small'', ``big'', ``large'', ...) or \textit{age} (``young'', ``old'', ...).
For each such class we fixed its meaning (e.g. \textit{evaluation}),
then came up with one or more adjectives for either side (e.g. ``good'', ``great'' vs ``bad'', ``stupid'').
This gave us an initial list of aspects with associated attributes (represented by adjectives) grouped into mutually exclusive sets.
We then iteratively expanded both sides of each aspect by using synonym and antonym information from thesaurus.
For example, for the aspect \textit{evaluation} thesaurus would be invoked to find synonyms of ``good'', add any of these synonyms to the left side of the aspect if they are mutually exclusive with all attributes on the right side, 
and add antonyms of ``good'' to the right side of the aspects if they are mutually exclusive with all attributes on the left side. 

Third, for each adjective-noun combination based on the noun list and the initial aspect list,
we used the Flickr-API (\url{https://www.flickr.com/services/api/}) 
to crawl images tagged with ``[adjective] [noun]'' (as one tag).
Finally, we iteratively performed the following steps for cleaning and structuring the data properly:
\begin{itemize}
    \item We removed all adjective-noun combinations with less than 20 occurrences.
    \item For each noun-aspect combination we counted the total number of available images 
    for each polarity. We only kept noun-aspect combinations if for each polarity the total number of available images was at least 100. 
    \item All nouns with less than 500 images in total and less than two available aspects were removed.
    \item We removed all aspects with a total of less than 500 images for any of the two polarities.
    \item We manually checked whether images obtained for individual adjective-noun combinations captured the associated aspect and visibly included the noun.
    All combinations where this was not the case were removed. Note that this led to the removal of almost half of the originally crawled data.
    \item For each feasible noun-aspect combination we randomly sampled the same number of images for left and right polarity from all relevant adjective-noun combinations.
    (Not recycling any images, i.e., each image in our dataset is only used for exactly one noun-aspect combination.)
\end{itemize}

\begin{table*}[t]
\begin{center}
    \begin{tabular}{|c|c|c|c|}
        \hline
        {\bfseries No.} & {\bfseries Aspect Name} & {\bfseries Attributes with Left Polarity (-1)} & {\bfseries Attributes with Right Polarity (+1)} \\
        \hline
        1 & evaluation &
          \begin{tabular}{@{}c@{}}[``perfect'', ``great'', ``good'', ``awesome'', \\``wonderful'', ``cool'', ``nice'']\end{tabular} 
          &
          [``bad'', ``stupid''] 
          \\
        \hline
        2 & size &
          [``small'', ``miniature'', ``little''] 
          &
          [``large'', ``big'', ``giant'', ``huge''] 
          \\
        \hline
        3 & age &
          [``new'', ``modern'', ``young''] 
          &
         \begin{tabular}{@{}c@{}} [``ancient'', ``old'', ``elderly'', ``historic'', \\``mature'', ``senior'', ``aged'']\end{tabular} 
          \\
        \hline
        4 & happiness &
          [``smiling'', ``laughing'', ``happy''] 
          &
          [``crying'', ``sad''] 
          \\
        \hline
		5 & rareness & 
          \begin{tabular}{@{}c@{}} [``exotic'', ``unusual'', ``strange'',  ``weird'',\\ ``odd'', ``peculiar''] \end{tabular}
          &
          [``normal'', ``everyday'', ``regular'', ``common''] 
          \\
        \hline
        6 & activity &
          [``active'', ``busy''] 
          &
          [``lazy'', ``tired'', ``sleepy'', ``sleeping''] 
          \\
         \hline
    \end{tabular}
\caption{Aspects in the \textit{aspects-DB} dataset. We only list attributes that are included in any adjective-noun combination in the dataset.}
\label{tab:aspects}
\end{center}
\end{table*}
The final \textit{aspects-DB} dataset contains $67,818$ images in total and features $13$  nouns for $6$ aspects.
A complete list of aspects can be found in Table \ref{tab:aspects}.
More detailed statistics are presented in Table~\ref{tab:dataset}.
The dataset is balanced on polarity level, i.e., for each noun-aspect combination, half of the the available images belong to the left polarity of the aspect and the other half to the right.
Since the ground truth was obtained by adjective-noun pairs, we keep the adjective part in our dataset as extra information,
so for each image, \textit{aspects-DB} includes
a noun, an aspect, the polarity of this aspect and the original adjective the noun was combined with in the adjective-noun tag.

\begin{table}
\begin{adjustbox}{max width=\columnwidth}
  \begin{tabular}{|c||c|c|c|c|c|c|}
    \hline
    \multirow{2}{*}{\bfseries Noun} & \multicolumn{6}{c|}{\bfseries Aspect} \\ 
      & {\bfseries eval.} & {\bfseries size} & {\bfseries age} & {\bfseries happ.} & {\bfseries rar.} & {\bfseries act.}  \\
    \hline \hline
	people &           0 & 0 & 7352 &   0 & 1620 &   0  \\
	guy &           1672 & 0 &  296 &   0 & 0 &   0 \\
	man &            \underline{554} & 0 &   5846 &   0 & 0 &   0 \\
	baby &            0 & 0 &    0 & 690 & 0 & 298  \\
	boy &          1558 & 0 &   0 & \underline{602} & 0 &   0 \\
	\hline
	cat &           874 & 0 &  960 & \underline{214} & 0 & 0 \\
	dog &          2094 & 0 &    \underline{402} & 630 & 0 & 922 \\
	\hline
	building &        0 & \underline{312} &  9912 &   0 & 0 & 0 \\
	house &           0 & 2084 &   7814 &   0 & 0 &  0 \\
	architecture &  528 & 0 &  8746 &   0 & 0 & 0 \\
	hotel &         \underline{342} & 0 &   5384 &   0 & 0 & 0 \\
	\hline
	city &             0 & 698 &   \underline{2372} &   0 & 0 & 286 \\
	\hline
	tree &            0 & 2428 &  328 &   0 &   0 & 0 \\
    \hline
  \end{tabular}
  \end{adjustbox}
  \caption{Numbers of images for all noun-aspect combinations in our final dataset.
  For all combinations that are withheld during training for 0-shot experiments (see Section~\ref{sec:tasks}), the corresponding numbers are underlined.}
  \label{tab:dataset}
\end{table}

Table~\ref{tab:samples} shows a few examples of nouns, their top aspects and the corresponding left and right polarities.
The dataset is available to the public and can be downloaded at \url{http://madm.dfki.de/downloads}.

We would like to emphasize that the ground truth in \textit{aspects-DB} is meant to capture general tendencies in subjective interpretation (where we use tags as proxy).
These tendencies must to some extent be corpus / domain specific and on item-level we cannot expect perfect performance.
This means that the task is not to detect objectively correct labels as in many common image classification datasets,
but to model general biases such as
``for this image of a sleepy puppy and noun \textit{dog}, people would typically interpret the image with respect to aspect \textit{age}. Aspects \textit{age}, \textit{activity}, \textit{evaluation} would likely be rated as having polarities - (\textit{young}), + (\textit{sleepy}), and - (\textit{good}) respectively''.

\section{Tasks} \label{sec:tasks}

\subsection{Aspect prediction} \label{sec:aspect_prediction}
In the first task, an image and a noun are given and the task is to predict which one of the aspects in our dataset (see Table~\ref{tab:aspects})
a subjective interpretation would most likely focus on.
For example, given an image with a puppy together with the noun ``dog'', a likely aspect from our list would typically be \textit{age}.
This problem is modeled as multi-class classification task, where for each given image and noun, only a single aspect is considered to be correct.

Note that our dataset is not balanced on aspect level,
and indeed, for a given noun the aspect prior typically strongly favors a certain aspect (see dataset statistics in Table~\ref{tab:dataset}). 
This skewness motivates us to not use an evaluation metric that is based on accuracy.
More precisely, for each noun we first compute the average F1 score across all applicable aspects,
and then average this number across all nouns to obtain the overall performance measure of a model:
$$ \mathtt{F1}_{\mathtt{asp}} := \frac{1}{|\mathcal{N}|} \sum_{n \in \mathcal{N}} \bigg( \frac{1}{|\mathcal{A}(n)|} \sum_{a \in \mathcal{A}(n)} \mathtt{F1}_n(a) \bigg), $$
where $\mathcal{A}(n)$ denotes all aspects that are available for noun $n$,
$\mathcal{N}$ the set of nouns, and $\mathtt{F1}_n(a)$ is the F1 score for aspect $a$ calculated over all images for noun $n$.

All available data is for each polarity split into 50\% training, 20\% development and 30\% test data.

\subsection{Aspect polarity detection}
\textit{Aspect polarity detection} is about deciding which polarity applies to a given noun for a given aspect in the context of the input image.
Coming back to the previous puppy example of Section~\ref{sec:aspect_prediction},
the true polarity for aspect \textit{age} would be ``left'' (corresponding to young age) when given an image of a puppy with the noun context ``dog''.
For training and evaluation we only consider one aspect at a time, hence this problem can be seen as binary classification task.

Apart from the \textit{standard} polarity detection task, where the same dataset split is used as for aspect prediction,
we consider \textit{0-shot} polarity detection,
where some aspect-noun combinations (the ones that are underlined in Table~\ref{tab:dataset}) are only contained in the final test set
and the rest of the data was randomly split into 70\% training and 30\% for development (per aspect-noun combination).
It should be noted that zero-shot learning on aspect prediction cannot be done in the same way (unless the noun is left out completely for training):
If we remove individual noun-aspect combinations and train a model on the remaining ones,
the model generally learns that for the nouns any excluded aspect is not feasible.
This points at another problem in the adjective-noun way of modeling, where aspect and aspect polarity are both blended into adjective information.


For calculating overall accuracy for polarity detection (for both sub-tasks),
for each aspect we compute the average accuracy across all nouns,
and then compute the average of these numbers:
$$ \mathtt{acc}_{\mathtt{pol}} := \frac{1}{|\mathcal{A}|} \sum_{a \in \mathcal{A}} \bigg( \frac{1}{|\mathcal{N}(a)|} \sum_{n \in \mathcal{N}(a)} \mathtt{acc}(a,n) \bigg), $$
where $\mathcal{A}$ denotes the set of aspects, $\mathcal{N}(a)$ returns a list of all nouns available for aspect $a$, and $\mathtt{acc}(a,n)$ denotes the accuracy for aspect $a$ and noun $n$.

\section{Methods} \label{sec:methods}
In this section we explain the methods we compare in our experiments (Section~\ref{sec:experiments}), 
where they are evaluated on both tasks described in the previous section.
For all models, except the XResNet variants, visual features are extracted from the image by an inception-v3 network \cite{inceptionv3}, which was trained on ImageNet \cite{imagenet} and kept unchanged.

\subsection{Logistic regression} \label{sec:log_reg}
We deploy various models based on logistic regression which take visual features from the inception network as the only input.
The motivation for using these models was to have a robust starting point which allows us to compare the effect of modeling the tasks in different ways.
More precisely, the different possibilities for modeling the tasks lead to these three versions (for both aspect and aspect polarity prediction):
\begin{itemize}
\item The \textit{noun-agnostic} version does not consider noun information at any point.
Aspect prediction is modeled as classification task with multiple classes.
So for predicting the most likely aspect given an image and noun, a single logistic regression model is trained to output the corresponding class from the visual features, irrespective of the noun.
Aspect polarity detection is modeled as separate binary classification problems,
i.e., for each aspect, one logistic regression model is trained to detect the polarity of the respective aspect from the image vector, again not taking the noun into account.
\item In the \textit{noun-specific} variant, separate models are trained for distinct nouns.
For each individual noun, we then follow the same approach as described in the previous point.
This means that for each noun we have one model predicting the most likely aspect, and for each noun-aspect combination we have one model for aspect polarity detection.
We explore this possibility as a simple way to take the noun context into account.
\item Finally, we consider a logistic regression model (\textit{adj-noun}) trained on detecting adjective-noun combinations from the inception features.
We include this model to analyze the effect of modeling the output as adjective-noun as compared to aspect and polarity.
Here, a single model is trained, and conditioning on a noun is done by simply ignoring all outputs with a different noun.
To evaluate this model on aspect prediction and aspect polarity prediction, the remaining adjective-noun scores have to be converted to aspect and polarity scores:
For aspect prediction, we select the adjective from the highest ranked adjective-noun combination and return the 
aspect it is contained in.
In case of aspect polarity detection, for any given aspect the adjective-noun outputs are filtered further such that all remaining adjectives are included in this aspect.
The polarity of the highest ranked among these adjectives is given as final polarity output.
\end{itemize}

In all these cases we use the scikit-learn \cite{scikit-learn} implementation for training and inference.

\begin{figure}[ht]
\begin{center}
   \includegraphics[width=\linewidth]{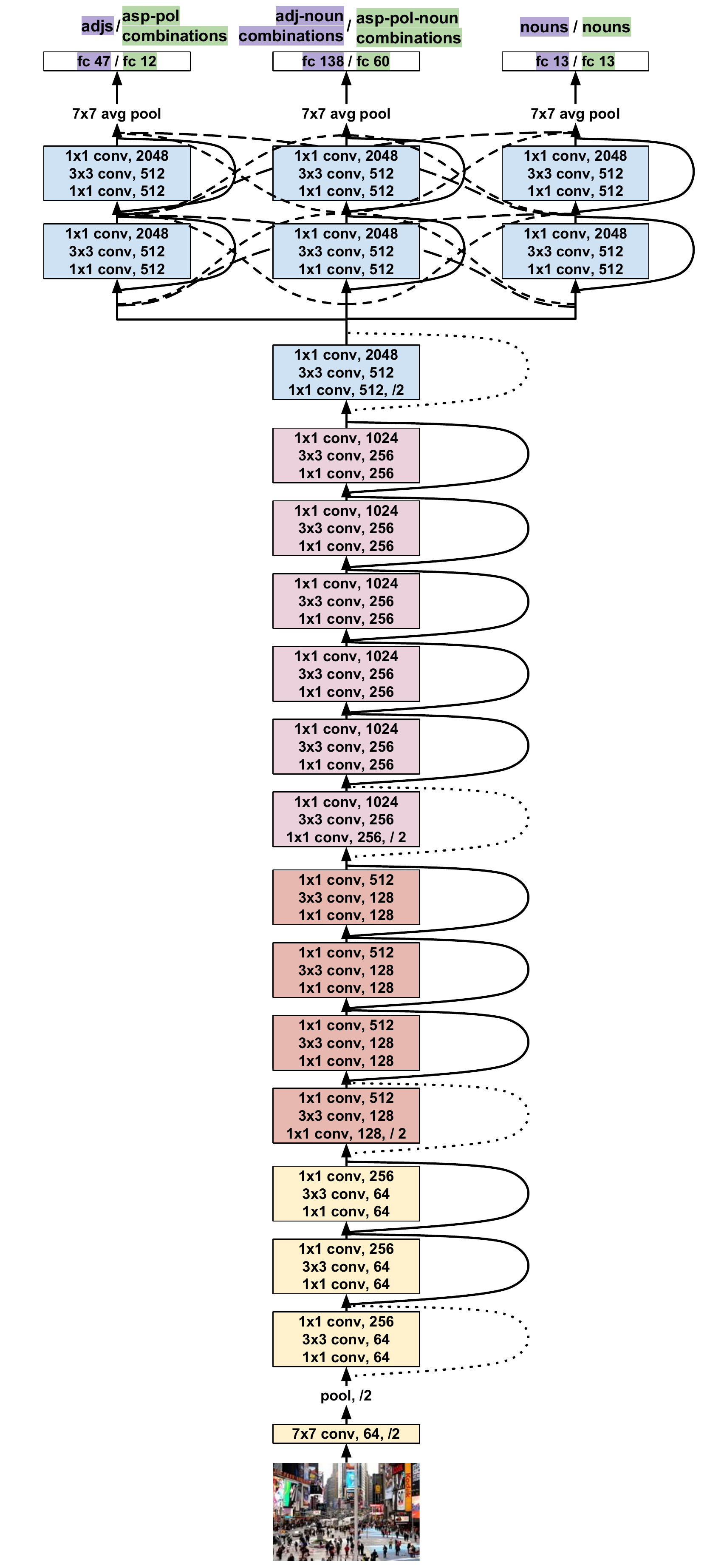}
\end{center}
   \caption{XResNet architecture (adapted from \cite{xresnet}). Solid shortcuts indicate identity, dotted connections indicate $1\times 1$ projections, and dashed shortcuts indicate cross-residual weighted connections.
   We train and evaluate two different versions of XResNet, one which predicts adjective, adjective-noun and noun, and one which predicts aspect-polarity, aspect-polarity-noun and noun (indicated in the diagram by purple and green color respectively).} 
\label{fig:xresnet}
\end{figure}

\subsection{Cross-residual networks} \label{sec:xresnet}
Cross-residual networks, or short XResNet, refers to an architecture which was introduced in \cite{xresnet} for adjective-noun pair detection
and is based on the well-known residual networks (ResNet) architecture \cite{resnet}.
Figure~\ref{fig:xresnet} shows the structure of the XResNet architecture we used.
The main difference of XResNet as compared to ResNet is that the network branches out at the end into three distinct heads,
where these branches remain closely connected to each other via so-called cross-residual connections.
The standard XResNet architecture has 50 layers and finally branches out to predict adjectives, nouns and adjective-noun pairs respectively.

We trained this standard model based on adjective and noun ground truth in our dataset,
starting from a pre-trained model and using settings as described in \cite{xresnet} for fine-tuning on our data. 
Since this method outputs scores for adjectives, nouns and adjective-noun combinations,
the output needs to be converted into aspect and aspect polarity information for evaluation.
We consider two ways of doing this conversion, each based on a different output branch of the model:
\begin{itemize}
\item Using the adjective output (\textit{adj}): To convert adjective scores to a prediction of the most likely aspect, adjectives are ordered by score and the aspect of the highest ranked adjective is output. 
For aspect polarity prediction, for any given aspect, all adjectives that are associated with the current aspect are ordered by score and the polarity of the highest ranked among these adjectives is taken as prediction of the model.
Note that this version completely ignores any given noun information.
\item Using the adjective-noun output (\textit{adj-noun}): For both aspect and aspect polarity prediction the output is first filtered based on the given noun, such that only adjective-noun combinations featuring this noun are kept. All remaining adjective-noun combinations only differ in their adjective, and can thus be treated like a list of adjective scores. To obtain the final output we then follow the same steps as described in the previous point.
\end{itemize}

To exclude the possibility that training on adjective and nouns while testing on aspects and polarities is adversarial to final performance,
we train another XResNet model which directly predicts aspect-polarity-noun combinations, aspect-polarity combinations and the noun, instead of adjective-noun, adjective and noun.
In this case, the output does not need to be converted, but we still have two possibilities for evaluation,
a noun-agnostic one using the aspect-polarity scores for the prediction (\textit{asp-pol}),
and one based on aspect-polarity-noun output (\textit{asp-pol-noun}) where conditioning on the noun is done by ignoring all irrelevant output scores.

\subsection{Concatenation + MLP}
The concatenation model is a straightforward application of information fusion,
where a one-hot encoding of the noun is appended to the image embedding obtained from the inception network.
This concatenated vector is then used as the input to a multi-layer perceptron (MLP) with one hidden layer.

We build one such model for aspect prediction and one for detecting aspect polarity.
For both aspect prediction and aspect polarity detection, the corresponding model has one output neuron per aspect.
Note however, that for polarity detection during training and testing we only consider the output of the unit corresponding to the aspect which is processed at the time.

\begin{figure*}[ht]
\begin{center}
   \includegraphics[width=1\linewidth]{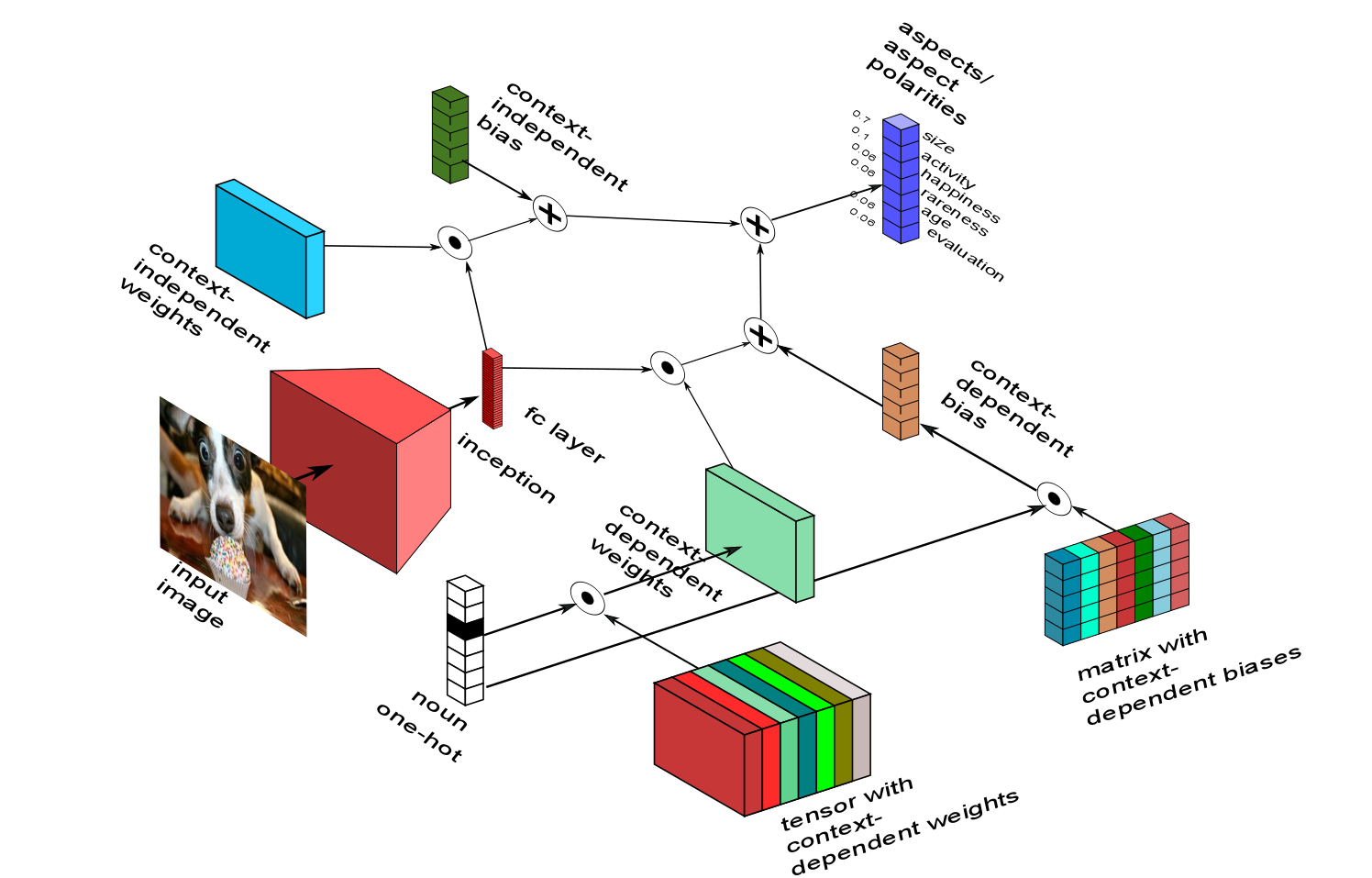}
\end{center}
   \caption{An overview of the Tensor Conditioning model. Given as input a one-hot encoded noun and an image, the Tensor Conditioning model embeds the image with a pre-trained inception-v3 network.
   This image embedding vector is then processed with the Tensor Conditioning layer which consists of two linear layers, a context-independent and a context-dependent one, followed by element-wise additive fusion of their outputs.
   For the context-dependent path, the Tensor Conditioning layer keeps a tensor with context-dependent weights and a matrix with context-dependent biases, which are multiplied by the noun vector to obtain weights and bias.
   (Since the noun is one-hot encoded, this multiplication amounts to a selection operation.)
   We use two separate Tensor Conditioning models for our experiments, one for aspect prediction with aspect likelihoods and one for aspect polarity detection with aspect polarities as output.}
\label{fig:custom_model}
\end{figure*}

\subsection{Tensor Conditioning}
Instead of merely concatenating image features and context we consider a slightly more sophisticated way of conditioning on the context,
using tensor products to combine information.
Similar ways of using context information have been used in several publications in the field of natural language processing,
for example \cite{nlp_anp}, \cite{Guevara:2010:RMA:1870516.1870521} and \cite{bamman},
but we are not aware of any other publication in computer vision including this method.

This approach is illustrated in Figure~\ref{fig:custom_model}.
The core part is the \textit{Tensor Conditioning layer},
which can be understood as part of a neural network that combines the noun-agnostic and noun-specific logistic regression models:
For each noun $i=1,\ldots,13$, there is a weight matrix $W_i$ and a bias term $b_i$.
In addition, the layer uses a weight matrix $W_0$ and a bias term $b_0$ that is used irrespective of the context.
Given as input the image embedding $x$ and the $i$-th noun, the output of the Tensor Conditioning layer is then computed as

$$ \tanh\big( (W_0 + W_i) \cdot x + b_0 + B_i \big). $$

We now represent nouns as one-hot vectors $n \in \mathbb{R}^{13}$ and put together all noun weight matrices $W_i$ into a third-order Tensor $W \in \mathbb{R}^{6 \times 2048 \times 13}$ and all noun biases $b_i$ into a bias matrix $B \in \mathbb{R}^{6 \times 13}$. 
The final layer function $T(x,n)$ can be formulated by using a tensor product between the noun context and a weight tensor to obtain the weight matrix for the given noun:
{\small
\begin{align*}
  T(x,n)
	&= \tanh\left( \Big(W_0 + \sum_{i=1}^{13} W_i \cdot n_i\Big) \cdot x + b_0 + \sum_{i=1}^{13} B_i \cdot n_i \right) \\
	&= \tanh\big( (W_0 + W \cdot n) \cdot x + b_0 + B \cdot n \big)
\end{align*}
}
\textbf{}
As in the concatenation approach, we deploy separate Tensor Conditioning models for the two tasks
of aspect prediction and aspect polarity detection.

\section{Results} \label{sec:experiments}
For both tasks described in Section~\ref{sec:tasks}, we ran experiments with all conditioning methods explained in Section~\ref{sec:methods}.
All hyper-parameters (learning rate, number of hidden units for the concatenation method)
were fine-tuned based on performances on training and development data (see Section~\ref{sec:tasks}).
We report performances on the test data.

\subsection{Aspect prediction}
All results for the aspect prediction task are listed in Table~\ref{tab:aspect_prediction_f1}.
As a statistical baseline, we include the score of a method which ignores the image and randomly predicts an aspect with the probability $P(aspect|noun)$,
based on general dataset statistics.

\begin{table}[ht]
\begin{center}
    \begin{tabular}{|c|c|}
        \hline
        {\bfseries Method} & {\bfseries Aspect-F1} \\ 
        \hline
        statistical baseline &
         0.47 \\
         \hline
        logistic regression (adj-noun) &
         \textbf{0.64} \\
        logistic regression (noun-agnostic) &
         0.48 \\
        logistic regression (noun-specific) &
         0.63 \\
         \hline
        XResNet \cite{xresnet} (adj) &
         0.59 \\
        XResNet \cite{xresnet} (adj-noun) &
         \textbf{0.64} \\
        XResNet (asp-pol) &
         0.54 \\
        XResNet (asp-pol-noun) &
         \textbf{0.64} \\
         \hline
        concatenation + MLP (10) & 
         0.48 \\
        concatenation + MLP (500) & 
         0.51 \\
        Tensor Conditioning & 
         0.62 \\
         \hline
    \end{tabular}
    \caption{Aspect prediction performances of all models. All methods except the baseline and the XResNet models use inception-v3 to embed the image.
    The number in parentheses after MLP indicates the number of hidden units used for this model.
    Please refer to Section~\ref{sec:methods} for details on the individual models.}
    \label{tab:aspect_prediction_f1}
    \end{center}
\end{table}

As we can see when comparing performances of the models to the statistical baseline,
noun-agnostic logistic regression and the concatenation model apparently did not work for this task at all.
Many of the other models achieve comparable scores, so there was no unique best performing method but a rather large group of top models.
Among these, Tensor Conditioning performs slightly worse than the rest (0.62 aspect-F1 while the top one achieves 0.64), but as expected performs very similarly to noun-specific logistic regression.
Interestingly, logistic regression (\textit{adj-noun}) is on par with the top XResNet models (\textit{adj-noun} and \textit{asp-pol-noun}), despite being a considerably simpler approach.

There are two methods between the top and bottom group: The \textit{asp-pol} and \textit{adj} versions of XResNet.
Here, surprisingly, predicting based on aspects and aspect polarities is worse than based on adjectives,
while this discrepancy disappears completely when using other output layers of the respective models (see \textit{adj-noun} and \textit{asp-pol-noun}).
In case of logistic regression the way of modeling also only marginally affects final performance.

Overall, the noun context seems to be important and is used by the models,
but this effect is less pronounced for the more sophisticated XResNet models.

\subsection{Aspect polarity detection}
Results for both polarity detection experiments can be found in Table~\ref{tab:polarity_accuracy}.

\begin{table}[ht] 
    \begin{tabular}{|c|c|c|c|c|c|}
        \hline
        \multirow{2}{*}{\bfseries Method} & \multicolumn{2}{c|}{\bfseries Polarity accuracy} \\ 
        & {\small\bfseries standard} & {\small\bfseries 0-shot} \\
        \hline
        statistical baseline &
         50.0\% & 50.0\% \\
         \hline
        logistic regression (adj-noun) &
         79.1\% & - \\
        logistic regression (noun-agnostic) &
         75.3\% & 63.4\% \\
        logistic regression (noun-specific) &
         \textbf{79.7\%} & - \\
         \hline
        XResNet \cite{xresnet} (adj) &
         75.5\% & 61.5\% \\
        XResNet \cite{xresnet} (adj-noun) &
         78.8\% & - \\
        XResNet (asp-pol) &
         76.4\% &  60.5\% \\
        XResNet (asp-pol-noun) &
         \textbf{79.7\%} & - \\
         \hline
        concatenation + MLP (10) & 
         69.1\% & \textbf{65.1\%} \\
        concatenation + MLP (500) & 
         71.4\% & 63.8\% \\
        Tensor Conditioning & 
         79.2\% &  61.8\% \\
         \hline
    \end{tabular}
    \caption{Aspect polarity detection performances of all models. All methods except the baseline and the XResNet models use inception-v3 to embed the image.
    The number in parentheses after MLP indicates the number of hidden units used for this model.
    Please refer to Section~\ref{sec:methods} for details on the individual models.
    Note that not all models are applicable to the 0-shot learning task.}
    \label{tab:polarity_accuracy}
\end{table}

The two best performing models for the \textit{standard} sub-task are noun-specific logistic regression and the \textit{asp-pol-noun} XResNet,
both of which are not applicable to 0-shot.
Among all models with 0-shot capability, Tensor Conditioning performs best on the standard task.
Again, the concatenation method works worst among all models, but this time at least it outperforms the baseline by a large margin.
For this task, modeling according to aspects and polarities seems to work better in general than using adjectives and nouns (see corresponding logistic regression or XResNet results).
As for aspect prediction, noun information is used but does not lead to terribly large improvements as compared to methods not using any noun context.

The performances on the 0-shot experiment are quite comparable,
while concatenation achieves the highest overall score.
This is an interesting finding, considering that it is worst for polarity detection in the standard task.

\section{Conclusion}\label{sec:conclusion}

We introduced a new method for capturing subjectivity prevalent in images.
To overcome several challenges, including the heavy bias towards positive tags / titles in social media, and to make it possible to separately evaluate different parts of subjective visual interpretation, we compiled a new dataset.
We ran our experiments on the new dataset and reported the results with different architectures.
It was also shown that with the new model, it is possible to perform \textit{0-shot} learning to predict unseen noun-attribute combinations.
Given the prevalence of simple concatenation for combining information in deep learning approaches, we find it interesting that Tensor Conditioning performed better in two out of three tasks.

Our results raise some fundamental questions, which we want to investigate in the future:
\begin{itemize}
    \item How can context be modeled optimally? Often researchers use concatenation as default choice and focus on data or hyper-parameters for improvement without changing this part of the architecture, but our results showed a decrease in performance in two out of three cases with the concatenation method.
    \item Which properties of the tasks make some methods (like concatenation) fail in one but outperform all other methods in another?
\end{itemize}

Furthermore, we plan to explore more ways of conditioning on context, and adapt our approach to applications such as personalized tag prediction and affective image captioning,
where biases at different stages of subjective visual interpretation according to the Focus-Aspect-Polarity model
can be made dependent on a user context to mimic the subjective interpretation of the given user.

\section*{Acknowledgments}
This work was supported by the BMBF project DeFuseNN (Grant 01IW17002) and the NVIDIA AI Lab (NVAIL) program.

{\small
\bibliographystyle{ieee}
\bibliography{egpaper}
}

\end{document}